\documentclass[journal]{IEEEtran}
\usepackage{amsmath}
\usepackage{lineno,hyperref}
\usepackage{booktabs}
\usepackage{multirow}
\usepackage{graphicx}
\usepackage{float}
\usepackage[justification=centering]{caption}
\ifCLASSINFOpdf
\else
\fi
\begin{document}
%
\title{Reinforcement Learning Based Emotional Editing Constraint Conversation Generation}

\author{Jia Li,~\IEEEmembership{Student Member,~IEEE,}
        Xiao Sun,~\IEEEmembership{Member,~IEEE,}
        Xing Wei,~\IEEEmembership{Member,~IEEE,}
        Changliang Li,~\IEEEmembership{Member,~IEEE, and}
        Jianhua Tao,~\IEEEmembership{Member,~IEEE}
\thanks{J. Li, X. Sun and X. Wei are with the school of Computer and Information, Hefei University of Technology,
 Hefei 230009, China. E-mail:lijiajia@mail.hfut.edu.cn; sunx@hfut.edu.cn; weixing@hfut.edu.cn}
\thanks{C. Li is currently the head with the Kingsoft institute of Artificial Intelligence. E-mail: lichangliang@kingsoft.com.}
\thanks{J. Tao is with the National Laboratory of Pattern Recognition, Institute of Automation, Chinese Academy of Sciences, Beijing 100190, China. E-mail: jhtao@nlpr.ia.ac.cn.}

}
\maketitle
\begin{abstract}
In recent years, the generation of conversation content based on deep neural networks has attracted many researchers. However, traditional neural language models tend to generate general replies, lacking logical and emotional factors. This paper proposes a conversation content generation model that combines reinforcement learning with emotional editing constraints to generate more meaningful and customizable emotional replies. The model divides the replies into three clauses based on pre-generated keywords and uses the emotional editor to further optimize the final reply. The model combines multi-task learning with multiple indicator rewards to comprehensively optimize the quality of replies. Experiments shows that our model can not only improve the fluency of the replies, but also significantly enhance the logical relevance and emotional relevance of the replies.
\end{abstract}


\ifCLASSOPTIONpeerreview
\begin{center} \bfseries EDICS Category: 3-BBND \end{center}
\fi
\IEEEpeerreviewmaketitle
\section{Introduction}

\IEEEPARstart{I}{n} recent years, with the development of artificial intelligence and robotics, affective computing has become increasingly critical in the research on human-computer interaction. Artificial inte[1][4]lligence with both emotion and intelligence has higher practical value and significance~\cite{CS:02,Mai:15}. To achieve accurate artificial intelligence, it is necessary to facilitate natural human-computer interactions that integrate intelligence and emotion.

In addition to visual, speech and other forms of expression, text is a basic and essential mode of emotion expression and is widely used in daily life. The emotional calculation of text includes text emotion recognition and emotional text generation. There are many works on text emotion recognition, and generating emotional text is very challenging. It is difficult to consider emotions naturally and coherently because we need to balance grammaticality and expression~\cite{Liang:16}.
The present emotional text generation considers the rule method and is a task-oriented application, limiting the domain adaptability and scalability of the model. In recent years, most research efforts are focused on improving the quality of conversational content (e.g., fluency, diversity) ~\cite{Liu:17,Gaurav:18} while ignoring the generation of fine-grained emotional factors in text. In~\cite{Zhou:17}, the researchers first introduced emotion into the neural network language model and proved that emotional sentences have better performance than sentences generated by the traditional models without considering emotion. Other researchers have used reinforcement learning methods to generate emotional text ~\cite{Mahipal:17,Thomas:17}. In~\cite{Ke:18}, the researchers used the reinforcement learning method to minimize penalty items, further strengthened the constraints on the text emotions, and enhanced the emotional factors in text.

There are two shortcomings in the past work. First, a whole sentence is generated from left to right, which is not entirely in line with habitual human natural language expression. This approach also limits the variety and fluency of the generated text~\cite{L:16,Michel:15}. Second, existing models are not able to fully consider the emotional elements contained in a conversation. The emotional of replies is uncontrollable, and in some cases, emotion is undetectable.

Given the above deficiencies, this paper proposes an emotional editing constraint conversation content generation model based on reinforcement learning. This paper makes the following contributions:
\begin{itemize}{
\item The proposed emotional editor can select the template sentence based on the topic and emotion and further optimize the generation of replies. The generated replies have more accurate emotion.
\item The proposed model comprehensively constrains the generation of replies from three aspects: coherence, topic, and emotion by introducing the reinforcement learning.
\item The proposed model introduces the multi-task learning method to enhance the model effect and learn the coherence, topic, and emotion of a reply so that these indicators can coordinate and constrain each other.
\item The experiments show that the proposed model is better than previous models that consider only one-sided factors, including the fluency of replies, the relevance of emotion and the relevance of the topic.
}\end{itemize}

\section{Related Work}
Recently, a sequence-to-sequence model based on sequence prediction problems, which can be applied to large-scale datasets~\cite{I:14}, has been widely used in machine translation~\cite{K:14} and conversation generation~\cite{Liu:16}. Later, a large number of variant models based on that were proposed, focusing on improving the quality of text in terms of grammar and sentence patterns, including increasing the diversity of generated text~\cite{Liang:16}, introducing additional prior knowledge to generate more meaningful text~\cite{Liu:17,Shuman:18}.

The work in~\cite{Partala:04,Prendinger:05} verifies that machines that can generate meaningful and emotional replies can enhance the users' satisfaction and lead to a smarter interaction.
However, in the above work, emotional factors are less considered in the text generation process. In~\cite{Zhou:17}, the researchers introduced the emotion category vector and two storage mechanisms to generate the replies of the corresponding emotions, the quality of the replies was improved compared with the past. In~\cite{Jingyuan:18}, the researchers introduced topic information and emotional information. Emotional keywords and topic keywords were predicted to guide the generation of replies so that the replies have higher topic relevance and emotional relevance.

Unsupervised text generation is also an important research field of natural language processing.  In~\cite{Jiwei:16}, the researchers combined the traditional sequence-to-sequence model with the reinforcement learning and proposed a model with information flow, semantic coherence, and ease of answer as the rewards, which improved the quality of text.
In addition, the generative adversarial network (GAN)~\cite{Goodfellow:14} is a novel unsupervised generation model that is similar in nature to reinforcement learning and has many applications in text generation. In~\cite{Lantao:17}, the researchers introduced the reinforcement learning to address the weakness that the GAN is indifferentiable to discrete sequence data. In~\cite{Ke:18}, the researchers used the Monte Carlo search to calculate the penalty term in the generation process and minimized the expectation of the overall penalty term as the objective function. The emotional constraint in the generation process was strengthened.

The above method uses the reinforcement learning strategy and the tradition neural network to generate the emotional conversation, but there are still two shortcomings: First, because natural language belongs to high-level semantic coding, it is difficult to find perfect objective indicators to measure it. Second, the works are unable to effectively excavate the emotional elements in conversation. The emotional strengths of generated replies are uncontrollable and inconspicuous. It is difficult to give full play to the role of emotion in conversation, and the resulting replies appear to be very blunt and rigid. The lexical, syntactic, grammatical and other information related to emotional factors is not considered.

\section{Emotional Conversation Generation Model}
This section discusses the proposed emotional conversation generation model in detail. We use $x$ to represent a post input by the external environment and $y$ to represent a reply given by the agent to the input. The components (states, actions, reward, etc.) of model are summarized in the following sub-sections. Due to the length limit, we provide the model details in the supplementary file.

\subsection{Action in Model}
An action is the process of generating a reply to an input post. The action space is infinite since arbitrary-length sequences can be generated.
\subsubsection{The Overview of Action}

Given post x, the encoder is utilized to obtain the encoded vector. After that, the process of generation consists of the following four steps:

Step I: The structure predictor is first used to predict whether an emotional keyword or topic keyword needs to be included in the reply and to predict the positional relationship between them.

Step II: Based on the result of Step I, a keyword predictor is used to generate corresponding keywords, and these keywords are used as prior knowledge to guide the generation of replies.

Step III: After the keywords are generated, the asynchronous decoder is used to generate the reply. The model considers two cases: when only one keyword exists, an asynchronous decoder similar to~\cite{L:16} is used to generate the reply. When the reply requires two keywords, the reply is divided into three clauses with the keywords as the boundary. The decoder generates these three clauses in turn. The three clauses are then combined into a complete reply according to the positional relationship.

Step IV: A template sentence is selected from the training set based on the emotional keyword and the topic keyword, and then, the template sentence is used to generate a corresponding emotion editing vector. The template and the emotion editing vector are used to edit and optimize the reply generated in Step III, thereby further improving the emotional accuracy and content quality of the reply.

\subsubsection{Keywords Predictor}
The main role of the keyword predictor is to predict which keywords should appear in the reply. The emotion dictionary and topic dictionary used are the same as the work in~\cite{Jingyuan:18}.
We first use the pre-trained LDA model to analyse the input post, predict the topic category of the reply, and determine the emotion category by artificially designating one of the seven categories listed above. To integrate the prior knowledge into the process, we combine the sum of hidden states  $\tilde{h}=\sum_{i=1}^Th_i$ and the category embedding $k=\{k^{et},k^{tp}\}$, keywords are predicted as follows:
\begin{equation}p(w_{et}^k|x,k^{et})=softmax(W_{et}^w[\tilde{h};k^{et}])\end{equation}
\begin{equation}p(w_{tp}^k|x,k^{tp})=softmax(W_{tp}^w[\tilde{h};k^{tp}])\end{equation}
where $w_{et}^k$ and $w_{tp}^k$ separately represent the emotion keyword and topic keyword that is expected to appear in the reply.

\subsubsection{Asynchronous Decoder}

\begin{figure}[h]
	\centering
	\includegraphics[width=9cm]{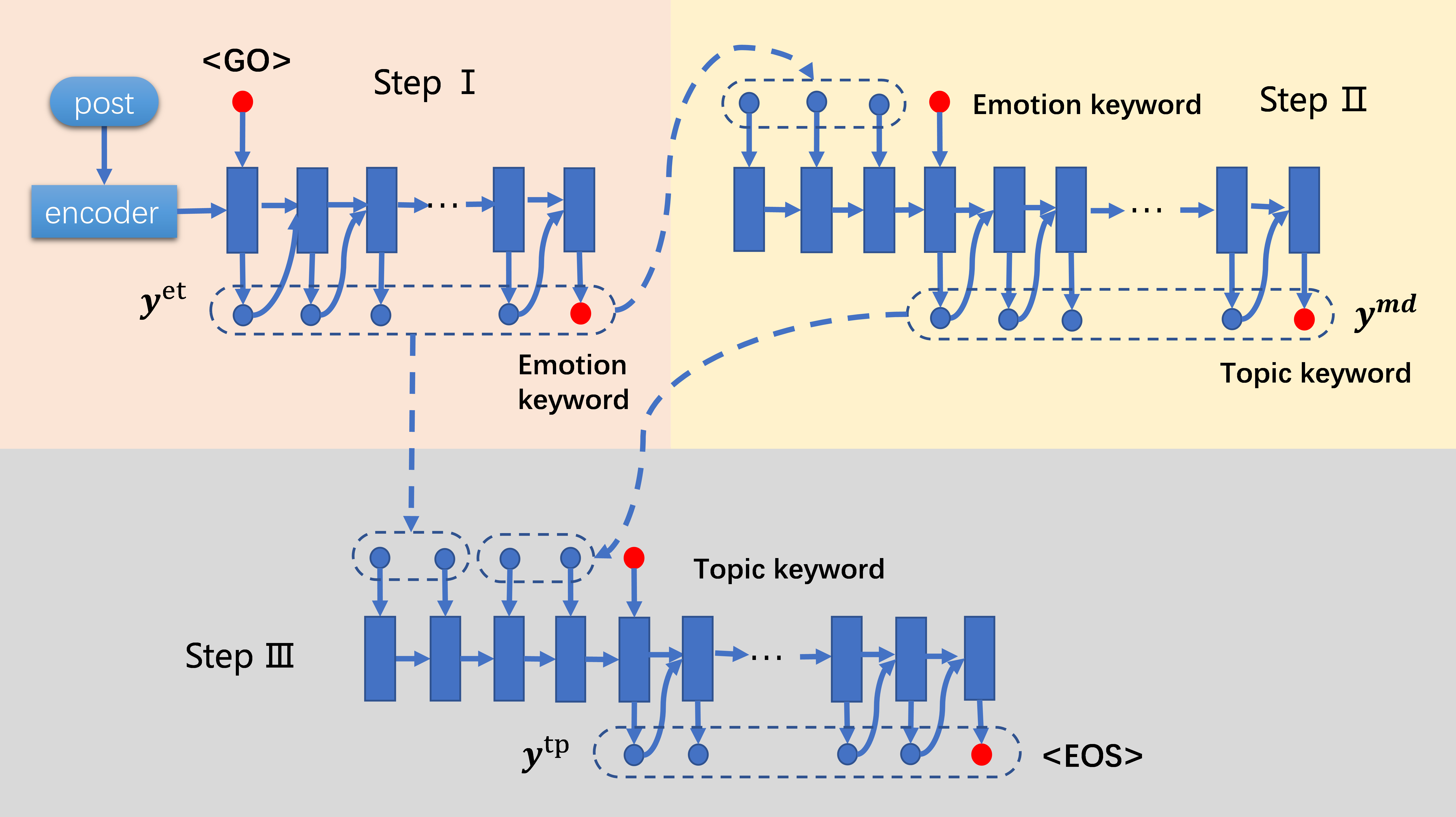}
	\caption{Asynchronous decoder structure}
\end{figure}

The next step is to generate the reply based on this prior knowledge. For the case where only one keyword is included, we use this keyword as the starting point and then go backward and forward to generate other parts of the reply. For the case where two keywords are included, because there are many situations, one of them is selected for detailed description to facilitate discussion. That is, the emotional keyword is in front, and the topic keyword is in the back. Other situations can be analogized.

Formally, suppose that the input post is $x=(x_{1},x_{2},...,x_{T})$, and the reply is $y=(w_s,y^{et},w_{et}^k,y^{md},w_{tp}^k,y^{tp},w_e)$, where $w_s$ and $w_e$ represent the start word $\langle
GO \rangle$ and the terminator $\langle EOS \rangle$, respectively, and $w_{et}^k$ and $w_{tp}^k$ represent the emotional keyword and the topic keyword, respectively. $y^{et}$ represents the portion between $\langle GO
\rangle$ and the emotional keyword, $y^{md}$ represents the portion between the emotional keyword and the topic keyword, and $y^{tp}$ represents the portion between the topic keyword and $\langle EOS \rangle$.

As shown in Fig.1, the entire reply is divided into three clauses. First, we generate $y^{et}$ with $\langle GO \rangle$ and $w_{et}^k$ as the starting word and the ending word, respectively. Second, based on the $y^{et}$, we generate $y^{md}$ with $w_{et}^k$ and $w_{tp}^k$ as the starting word and the ending word, respectively. Third, based on the $y^{et}$ and the $y^{md}$ , $y^{tp}$ is generated starting from $w_{tp}^k$ to $\langle EOS \rangle$.
Then, the clauses and keywords are combined in the previously determined order to get a complete reply. The specific process is as follows:
\begin{equation}p(y^{et}|x,w_1^{k})=\prod_{i=1}^Mp(y_i^{et}|y_{i-1}^{et},s_i^{et})\end{equation}
\begin{equation}p(y^{md}|y^{et},w_2^{k})=\prod_{i=1}^Lp(y_i^{md}|y_{i-1}^{md},s_i^{md})\end{equation}
\begin{equation}p(y^{tp}|y^{et},y^{md},w_3^{k})=\prod_{i=1}^Np(y_i^{tp}|y_{i-1}^{tp},s_i^{tp})\end{equation}

where $w_i^k\in \{(w_s,w_{et}^k),(w_{et}^k,w_{tp}^k),(w_{tp}^k,w_e)\}$ denotes the set of keywords, $s_i^{md}$,$s_i^{tp}$ and $s_j^{tp}$ denote intermediate states in the decoding process of the three clauses, respectively.

\subsubsection{Emotional Editor}
\begin{figure}[t]
\centering
\includegraphics[width=9cm]{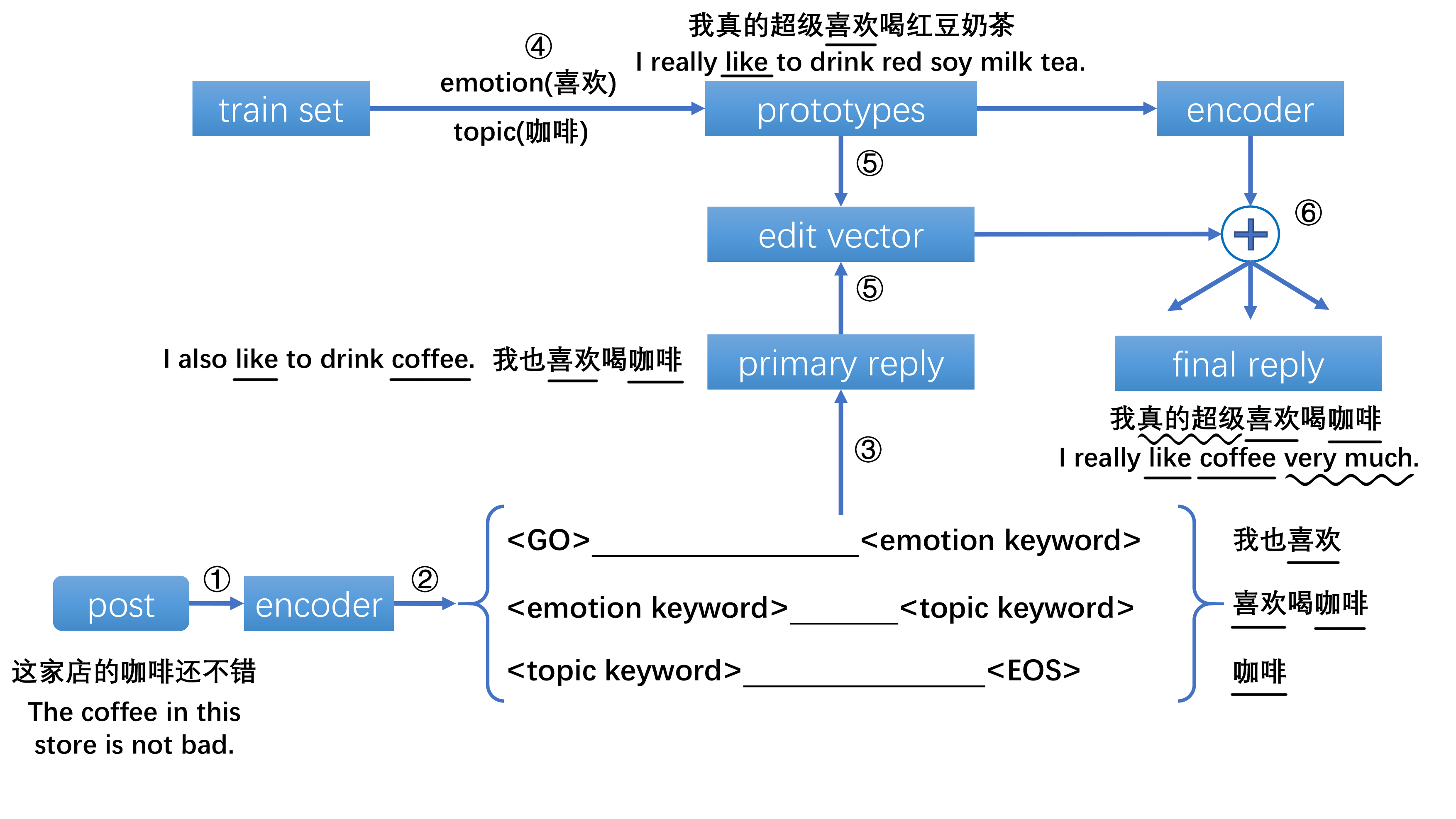}
\caption{Emotional content editing optimization process.}
\end{figure}

\textbf{(1) Picking a template:}
As shown in Fig.2, the words with single underlines denote the keywords and the words with wavy underlines denote the emotional editing part. We select the template $y^{'}$ in training set $\chi$ based on the keywords and the positional relationship of the keywords. The priority when selecting a template is as follows (decreasing): a sentence containing the same keywords and the same positional relationship, a sentence containing the same keywords but different positional relationships, a sentence containing only the same topic keyword, and a sentence containing only the same emotional keyword.
Using lexical-level similarities to distinguish between sentences with the same priority:
\begin{equation}L(y,y^{'})=d_J(y,y^{'})\end{equation}
Where $d_J(y,y^{'})$ is the Jaccard distance between template sentence $y^{'}$ and primary reply $y$. According to the above rules, the sentence with the highest priority and the highest similarity with the candidate reply is selected as the template sentence $y^{'}$.

\textbf{(2) Calculate the emotion editing vector:}
In ~\cite{Kelvin:18}, the authors suppose that multi-word insertions and deletions to be represented as the sum of the inserted word vectors. In contrast to the above work, to enhance the optimization effect of the edit vector on emotion, we introduce the emotion coefficient for each word in a sentence. The smaller the distance from the emotional keyword, the greater the emotional coefficient of the word is. We multiply the word vector of each word to be modified and the word emotion coefficient and then sum them, thereby calculating the emotion editing vector.
Formally, define $I=y/y^{'}$ to be the set of words added and $D=y^{'}\backslash$$y$ to be the words deleted. We represent the difference between $y^{'}$ and $y$ using the following vector:
\begin{equation}\alpha_w=\frac{1}{\sqrt{2\pi}\sigma}exp(-\frac{(l_w-\mu)^2}{2\sigma^2})\end{equation}
\begin{equation}f(y,y^{'})=\sum_{w\in I}\alpha_w\Phi(w)\bigoplus\sum_{w\in D}\alpha_w\Phi(w)\end{equation}
where $l_w$ represents the distance between word $w$ and emotional keyword and $\alpha_w$ represents the emotional coefficient of word $w$. $\Phi(w)$ represents the word vector of word $w$ and $\bigoplus$ represents a join operation.

Referring to Kelvin's work, we design $q$ to add noise to perturb the direction of vector $f$. We let $f_{norm}=\|f\|$,$f_{dir}=f/f_{norm}$ and let $vMF(v;\mu,\kappa)$ denote a $vMF$ distribution over points v on the unit sphere with mean vector $\mu$ and concentration parameter $\kappa$. Define the following:
\begin{equation}q(z_{dir}|y^{'},y)=vMF(z_{dir};f_{dir},\kappa)\end{equation}
\begin{equation}q(z_{norm}|y^{'},y)=Unif(z_{norm};[{\widetilde f_{norm}},{\widetilde f_{norm}}+\varepsilon])\end{equation}
where ${\widetilde f_{norm}}
=\min({\widetilde f_{norm}},10-\varepsilon)$ is the truncated norm. The resulting edit vector is $z=z_{dir} \cdot z_{norm}$.

\textbf{(3) Edit optimization}:We employ an encoder-decoder architecture to implement emotional editor, where prototype $y^{'}$ is the input sequence and revised sentence $y$ is the output sequence, extending it to condition on an edit vector $z$ by concatenating $z$ to the input of the decoder at each time step:
\begin{equation}p(y|y^{'},z)=\prod_{j=1}^Kp(y_i|y_{i-1},[s_i,z])\end{equation}
The emotional editing optimization of the reply is completed, and final reply $y$ is obtained.

\subsection{Rewards Calculation}
For the contents of state and policy, we refer to the work in~\cite{Jiwei:16}. The state is represented by post $x$ input by the external environment. Note that we use a stochastic representation of the policy (a probability distribution over actions given states). In this subsection, we discuss major factors that contribute to the success of a reply and describe how approximations to these factors can be operationalized in computable reward functions.

\textbf{Coherence}:$p_{seq2seq}(y|x)$ denotes the probability of generating reply $y$ given post $x$. $p_{seq2seq}^{backward}$ denotes the backward probability of generating post $x$ based on reply $y$. $p_{seq2seq}^{backward}$ is trained in a similar way as standard sequence-to-sequence models with sources and targets swapped. Again, to control the influence of replies length, both $\log p_{seq2seq}(y|x)$ and $\log p_{seq2seq}^{backward}(x|y)$ are scaled by the length of replies. $N_y$ and $N_x$ represent the length of the reply and the post, respectively. We calculate the coherence of reply $y$ with the following:
\begin{equation}
r_1=\frac{1}{N_y}\log p_{seq2seq}(y|x)+\frac{1}{N_x}\log
p_{seq2seq}^{backward}(x|y)
\end{equation}

\textbf{Topic relevance}:We use the pre-trained LDA model mentioned earlier to make topic category predictions for the reply. We define $k^{tp}$ as the topic category for the post, $LDA(y)$ as the predicted probability distribution of the topic of the LDA model for the reply, and $N_{tp}$ as the total number of topic categories. The topic relevance of reply $y$ is calculated by the following:
\begin{equation}
r_2=-\sum_{i=1}^{N_{tp}}k_i^{tp}\log(LDA_i(y))
\end{equation}

\textbf{Emotion relevance}:We use a convolutional neural network to classify the reply into sentiment categories and, based on the predictions, to see if the reply meets the pre-required sentiment categories. We define $k^{et}$ as the specified sentiment category, $D^{et}(y)$ as the predicted probability distribution of the classifier, and $N_{et}$ as the total number of sentiment categories. We calculate the emotional relevance of reply $y$ by the following:
\begin{equation}
r_3=-\sum_{i=1}^{N_{et}}k_i^{et}\log(D_i^{et}(y))
\end{equation}

To strengthen the constraints on the reply generation process, rewards are calculated for each clause, that is, the weighted sum of the indicators proposed above. Each clause has a different focus, so its weights in reward calculations are different. After repeated experiments, when we use the following weight parameters, the model has the best fitting effect on the corpus. The reward calculation formulas are organized as follows:
\begin{equation}
r^{et}=0.2r_1+0.2r_2+0.6r_3
\end{equation}
\begin{equation}
r^{md}=0.2r_1+0.4r_2+0.4r_3
\end{equation}
\begin{equation}
r^{tp}=0.2r_1+0.6r_2+0.2r_3
\end{equation}
\begin{equation}
r=0.5r_1+0.25r_2+0.25r_3
\end{equation}
where $r^{et}$, $r^{md}$ and $r^{tp}$ represent the rewards of the three clauses $y^{et}$,$y^{md}$ and $y^{tp}$, respectively, and $r$ represents the reward of the reply that is spliced and edited. The process of calculating the reward is shown in Figure 3. Therefore, the final reward $R$ for generating a reply is
\begin{equation}
R(a,[x,y])=r^{et}+r^{md}+r^{tp}+r
\end{equation}
\begin{figure}[h]
\centering
\includegraphics[width=8.5cm]{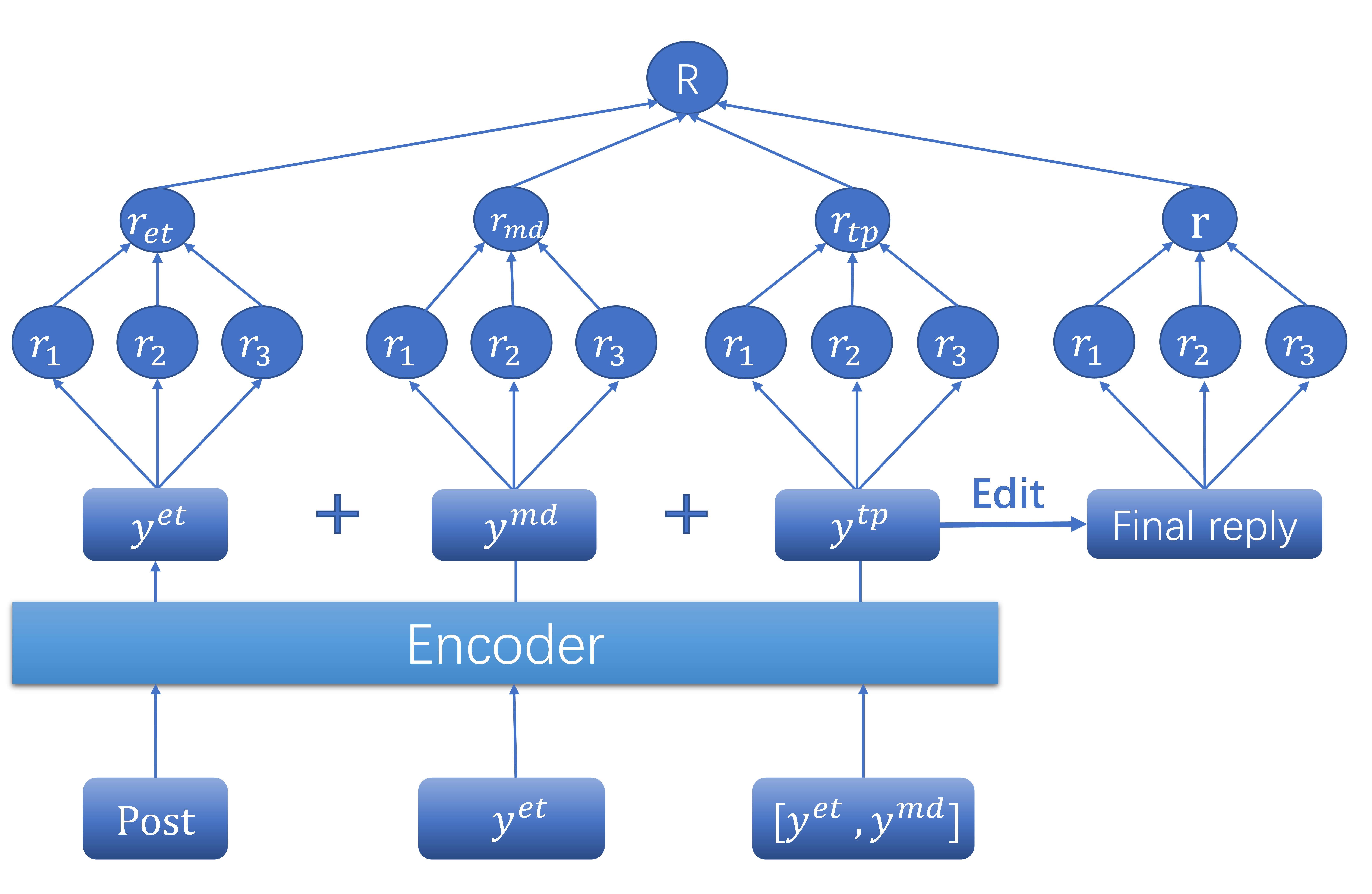}
\caption{Reward calculation process.}
\end{figure}
where $r^{et},r^{md}$ and $r^{tp}$ represent the rewards of the three clauses $y^{et},y^{md}$ and $y^{tp}$, respectively, and $r$ represents the reward of the reply that is spliced and edited. The process of calculating the reward is shown in Fig. 3. Therefore, the final reward $R$ for generating a reply is
\begin{equation}
R(a,[x,y])=r^{et}+r^{md}+r^{tp}+r
\end{equation}
The model uses multiple indicators to comprehensively consider the reply; therefore, to promote learning between indicators, the model introduces a multi-task learning strategy based on parameter sharing~\cite{Yu:18}. In the process of generating the reply, the encoder is shared, especially in the process of generating each clause. By using the same encoder, the indicators can be combined with each other, which is more conducive to measuring the quality of reply from the overall perspective.
\subsection{Optimization}
The model is able to generate some plausible replies by initializing the MLE parameters. We then use policy gradient methods to find parameters that lead to a larger expected reward. The objective to maximize is the expected future reward:
\begin{equation}
J_{RL}(\theta)=E_{p(a_1:T)}[\sum_{i=1}^TR(a_i,[x_i,y_i])]
\end{equation}
where $R(a_i,[x_i,y_i])$ denotes the reward resulting from action $a_i$. We use the likelihood ratio trick~\cite{Ronald:92} for gradient updates:
\begin{equation}
\nabla J_{RL}(\theta)\approx\sum_i\nabla\log p(a_i|x_i,y_i) \\
\sum_{i=1}^TR(a_i,[x_i,y_i])
\end{equation}
\section{Experiments and Results}
\subsection{Dataset Description}
The experiments used the emotional conversation dataset NLPCC2017 to train and test the proposed model. There are 1,118,341 post-reply pairs after the entire dataset has been filtered to remove meaningless sentences. Approximately 43.5\% of the conversation replies contain two keywords. Here, we focus on data experiments with two keywords.  A total of 80,000 pairs of sentences were randomly selected from the training set to train the LDA model, and 100,000 pairs of sentences were randomly selected to train the emotional classifier. Due to the length limit, we provide the implementation details in the supplementary file.
\subsection{Baseline Models}
In the experiments, our model is compared with the following baselines:

\textbf{Seq2Seq}: An encoder and a decoder for text generation can be used to generate some fluent text. We compare it to the quality of the generated text.

\textbf{ECM}: It introduced emotional embedding vectors and two stored mechanisms to generate emotional replies. We contrast it with the emotional intensity and emotional accuracy of the replies.

\textbf{SentiGAN}: Using GAN and the reinforcement learning strategy support the generation of emotional text. We contrast it with the emotional intensity and emotional accuracy of the replies.

\textbf{E-SCBA}: The model introduced both emotion and topic knowledge into the generation to make a comprehensive optimization for the quality of replies. We contrast it with the content quality and sentiment of the text.

\textbf{W/O Edit}: To verify the effect of the proposed emotion editor, the emotion editor is removed and compared to the complete model.
\subsection{Manual Evaluation}
We asked four annotators to evaluate the results of our model and baselines. In total, we used 700 conversations, 100 for each emotion category, which were sampled randomly from the test set. The annotators were asked to score a reply based on the following metrics:

Consistency measures the fluency and grammaticality of the reply on a three-point scale: 0, 1, 2.
Logic measures the degree to which the post and the reply logically match on a three-point scale as above. Note that overly short or overly frequent replies would be annotated as either 0 or 1 (if the annotator thought the reply related to the post), such as "Me too".
Emotion measures whether the reply includes the right emotion. A score of 0 means the emotion in the reply is wrong or there is no emotion, a score of 1 means the reply has the correct emotion but the intensity is weak, and a score of 2 means the reply has the correct emotion and the intensity is strong. Because SentiGAN is generally used for emotional text generation, no logical comparison is involved.

\begin{table*}
	\centering
	\vskip5mm \vbox{\vskip2mm
		\renewcommand{\baselinestretch}{1.2}
		{\footnotesize\centerline{\tabcolsep=4pt
				\begin{tabular}{c|c|c|c|c|c|c|c|c|c|c|c|c}
					\toprule
					{{Model}} & \multicolumn{3}{|c|}{{Overall}} &\multicolumn{3}{|c|}{{Happy}} &\multicolumn{3}{|c|}{{Sad}} &\multicolumn{3}{|c}{{Surprise}} \\
					\cline{2-13}
					&{{C}}&{{L}}&{{E}}   &{{C}}&{{L}}&{{E}}&{{C}}&{{L}}
					&{{E}}&{{C}}&{{L}}&{{E}}     \\
					\hline
					{{Seq2Seq}}      &1.288   &0.764   &0.430   &1.299 & 0.924&0.571 &1.384&0.928 &0.481 &1.182 &0.723 &0.152 \\
					{{SentiGAN}}     &1.347   &  -   &1.068   &1.425 &- & 1.285&1.500 &- &1.167 &1.200 &- &0.687   \\
					{{E-SCBA}}       &1.335   &1.122   &0.955   &1.421 & 1.286&1.230 &1.495 &1.267 &1.050 &1.197 &{\bf {0.901}} &0.500   \\
					{{W/O Edit}}     &1.336   &1.136   &0.994  &1.426 &1.285 &1.219 &1.504 &1.255 &1.101 &1.194 &0.897 &0.614   \\
					{{Ours}}         &{\textbf{1.390}}   &{\textbf{1.170}}   &{\textbf{1.135}}  &{\textbf{1.484}} &{\textbf{1.302}} &{\textbf{1.316}} &{\textbf{1.548}} &{\textbf{1.349}} &{\textbf{1.174}} &{\textbf{1.214}} &0.900& {\textbf{0.742}}  \\
					{{Ground Truth}} &1.739   &1.615   &1.312   &1.867 &1.728 &1.562 &1.808 &1.547 &1.186 &1.782 &1.627 &1.074  \\
					\bottomrule
	\end{tabular}}}}
	\caption{The results of manual evaluation (C = Consistency, L = Logic, E = Emotion).}
\end{table*}

Table \uppercase\expandafter{\romannumeral1} (2-tailed t-test: $p<0.05$ for Consistency and Logic, $p<0.01$ for Emotion) compares our model with the baselines. As we can see, the average performance of our model on the three indicators is better than that of other models. The experimental results are further analysed below.

Considering consistency and emotional relevance, our model is much better than others. However, the model without the emotion editor is not outstanding in relation to these two
indicators; in fact. This shows that the proposed emotion editor can improve the fluency and emotional relevance of the reply.
In terms of logic, our model does not achieve the best results for the surprised and angry emotions. This is mainly because the datasets for the two emotion categories are relatively small. This leads to the selection of a template sentence ignoring the constraints on the topic and the deviation of the optimized reply in the topic.

\begin{table}
\vskip5mm \vbox{\vskip2mm
\renewcommand{\baselinestretch}{1.35}
{\footnotesize\centerline{\tabcolsep=11pt
\begin{tabular}{c|c|c|c|c|c}
\toprule
{{Model(\%)}} & {2-2} & {1-2} & {1-1} & {1-0} & {0-1}  \\
\hline
{{Seq2Seq}}    &10.5&5.4 & 15.1 & 42.6 & 10       \\
{{SentiGAN}}   &24.8&21.7 & 36.9 & 5.6 & 11.3       \\
{{E-SCBA}}     &28.6&15.8 & 30 & 16.9 & 3.3        \\
{{W/O Edit}}   &27.5&14.9 & 31.7 & 15 & 9.7         \\
{{Ours}}    &\bf{41.7}&\bf{25.6} & 20.4 & 1.1 & 4.4    \\
\bottomrule
\end{tabular}}}}
\caption{Logical and sentiment scores in the manual assessment.}
\end{table}

The score distribution of each model in terms of logic and sentiment is calculated, as shown in Table \uppercase\expandafter{\romannumeral2}. For instance, 2-1 means logic score is 2 and emotion score is 1. As observed, the baseline models have a small proportion of 2-2, which indicates that they can't balance the emotion and the topic. However, the model proposed in this paper performs well in this respect, with the proportion of 2-2 reaching 41.7\% and the percentage of the emotional score of 2 reaching 67.3\%, which shows that the proposed model makes up for the shortcomings of the previous model's weak emotion.
\subsection{Automatic Evaluation}
We adopted perplexity to evaluate the model at the content level (to determine whether the content is relevant and grammatical). To evaluate the model at the emotion level, we adopted emotion accuracy as a reflection of agreement between the expected emotion category (as input to the model) and the predicted emotion category of a reply generated by the emotion classifier.
\begin{table}[h]
\centering
\vskip5mm \vbox{ \vskip2mm
\renewcommand{\baselinestretch}{1.35}
{\footnotesize\centerline{\tabcolsep=8pt
\begin{tabular}{c|c|c|c|c|c|c}
\toprule
  {{Model}} & \multicolumn{3}{|c|}{{Perplexity}} &\multicolumn{3}{|c}{{Accuracy}} \\
\cline{2-7}
       &{{1}}&{{2}}&{{3}}   &{{1}}&{{2}}&{{3}}\\
\hline
{{Seq2Seq}}      &67.4   &69     &68     &0.164 &0.188 &0.175 \\
{{SentiGAN}}     &65.1   &69.6   &66.7   &0.768 &0.790 &0.792 \\
{{E-SCBA}}       &64.8   &65.9   &66.1   &0.774 &0.769 &0.772 \\
{{W/O Edit}}     &65     &66     &66.5   &0.775 &0.759 &0.776 \\
{{Ours}}         &\textbf{62.2}   &{\textbf{61}}  &\textbf{61.4}  &{\textbf{0.871}} &\textbf{0.870} &{\textbf{0.869}} \\
\bottomrule
\end{tabular}}}}
\caption{The results of objective evaluation.}
\end{table}
The results of the experiment are shown in Table \uppercase\expandafter{\romannumeral3}. To avoid the contingency of the experiment, we performed 3 tests for each model. The results show that our model achieves the best results in terms of perplexity and emotional accuracy.
Compared with the model without emotion editor, the results show that the latter does not perform well in terms of perplexity and emotional accuracy. However, after the emotion editor is added, the model's performance greatly improves. This shows that the proposed emotion editor can integrate the keywords prior knowledge into the reply naturally. Smoothing, optimizing, and other editing operations are performed on the replies according to the template, which not only makes the reply more fluent but also makes the emotions of reply more prominent.
\begin{figure}[h]
\centering
\includegraphics[width=8cm]{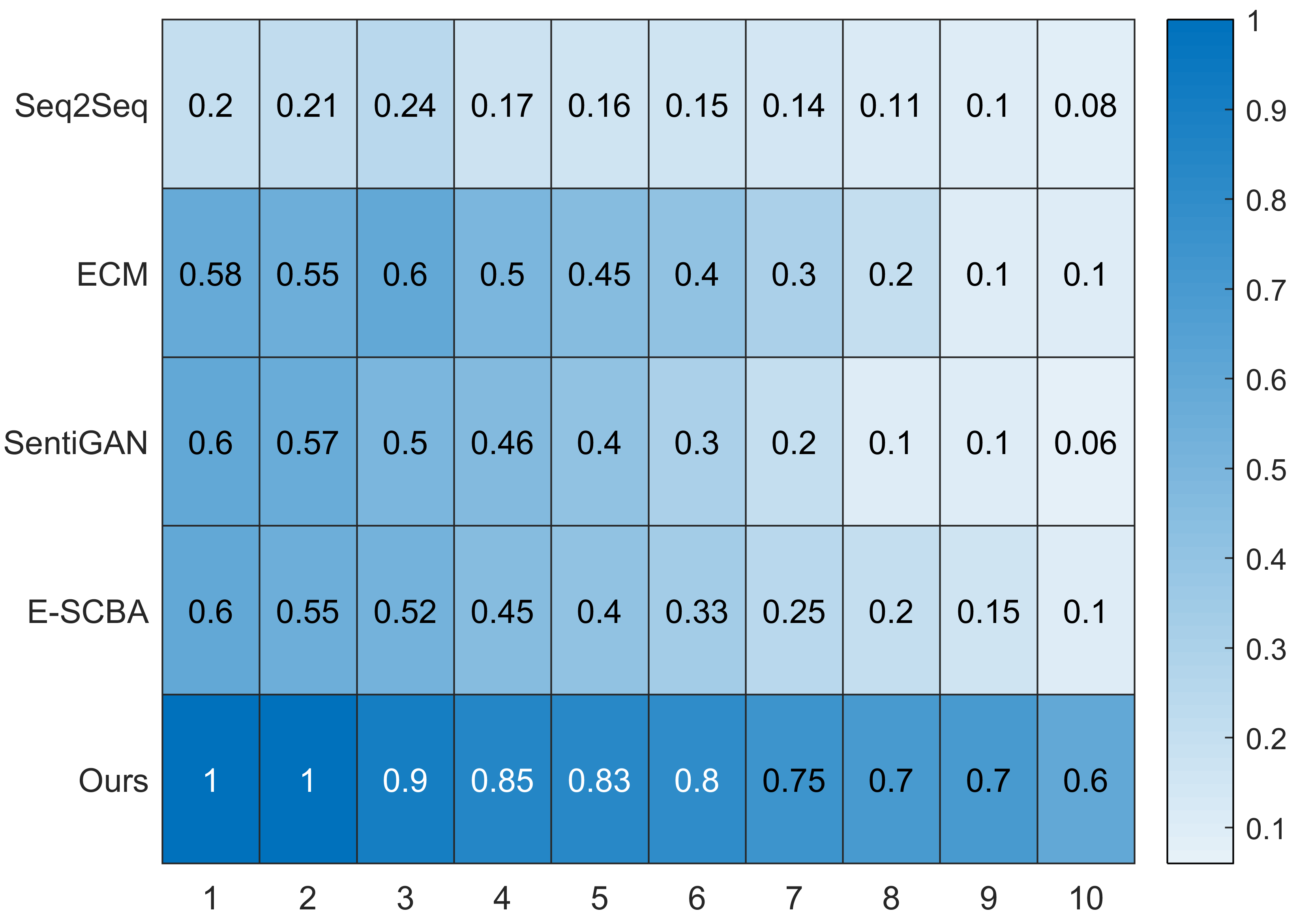}
\caption{The visualization of word distribution, where positions with deeper colour have a higher diversity.}
\end{figure}

In Fig.4, we visualize the diversity distribution of words in different positions (1-10) of the reply. Our model is committed to solving the problem of generic replies, which can be defined as a high frequency of certain replies to posts as well as a large number of identical words produced in the same place. The results shown in the figure have been normalized.

The worst model of all is the general Seq2Seq, whose diversity in different locations is always low. In addition to insufficient information from posts, the immutable sequential structure limits the potential of the model, resulting in generic replies. In contrast, our model obtains sufficient information in the process of decoding by introducing prior knowledge of keywords. The editor is then used to optimize replies to further improve the text quality and emotional relevance without generating a single secure reply. Besides, the colour of our model fades more slowly, showing that our model improves not only the quality of content but also the capacity of memory.

\begin{figure*}[h]
	\centering
	\includegraphics[width=13cm]{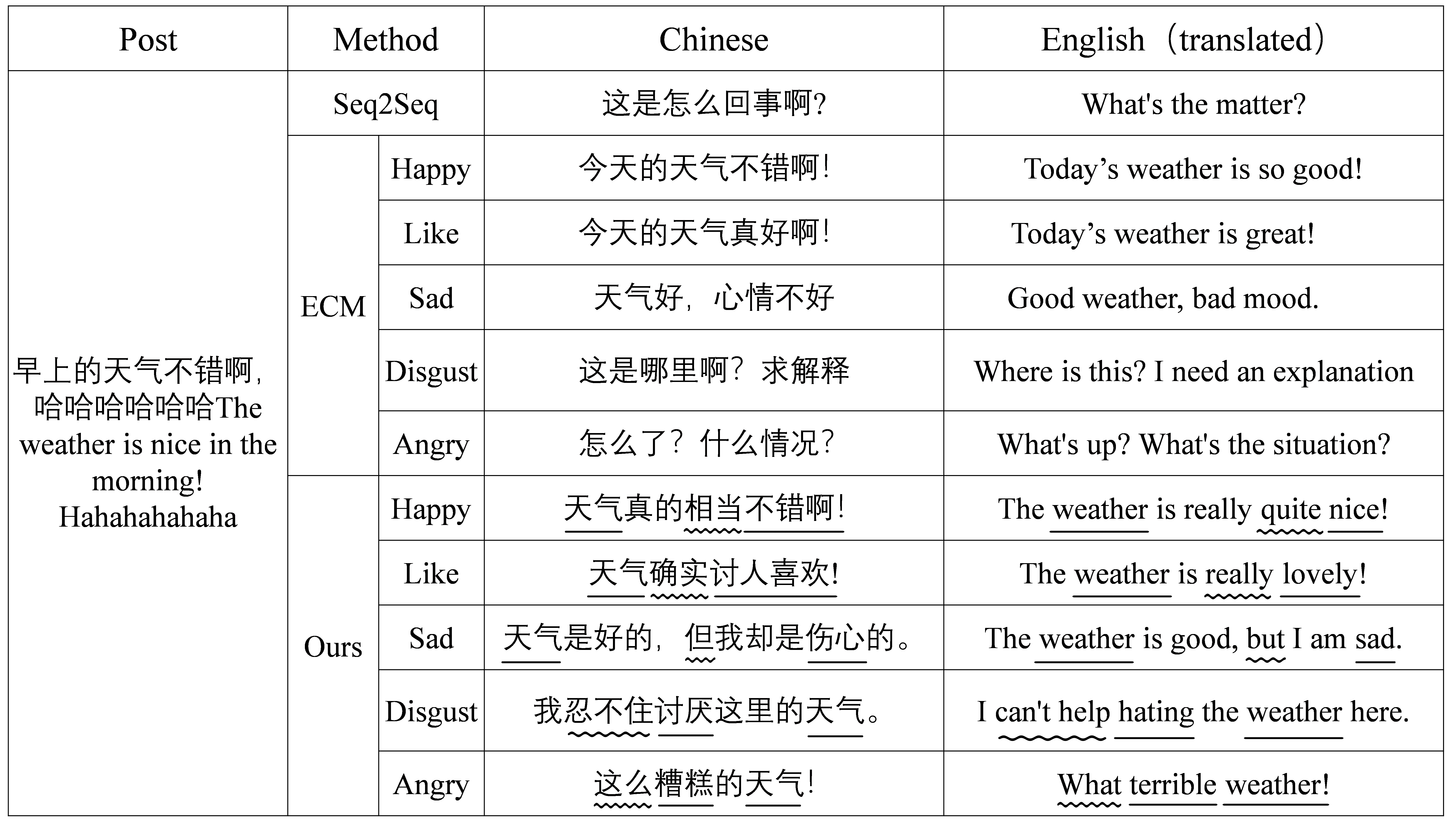}
	\caption{Sampled conversations with different emotions from the test data.}
\end{figure*}

\subsection{Case Study}
We provide some examples in Fig.5. The words with single underlines denote the keywords, and the words with wavy underlines denote the emotional editing optimization part. As we can see, the general Seq2Seq prefers to generate short and meaningless replies. The replies are more like a summary of the posts rather than a conversation.

The ECM model is improved compared to the Seq2Seq model, and it can generate fluent and emotional replies. However, the examples demonstrate that the most of the replies generated by the ECM have weak or even no emotion, such as "Where is it here? Ask for explanation", etc. This is because ECM only introduces the emotion embedding vector to guide the model to generate an emotional reply, which lacks detailed emotional guidance information when generated, resulting in fuzzy emotional replies. In contrast, the replies generated by our model, after the optimization of emotional editing, not only have rich and varied sentence patterns but also greatly enhance the emotional relevance and intensity. For example, the happy reply "I have never seen such a cute cake!" and the angry reply "Such bad weather!" contain more realistic and more detailed emotions.

\subsection{Conclusions and Future Work}
This paper proposes an emotional conversation generation model based on the reinforcement learning. Our model generates replies in three iterations and proposes a mechanism of emotional editing that refers to existing sentences to further strengthen the content quality and emotional relevance of the replies. Subjective and objective experiments show that the model proposed in this paper can generate logical and emotional replies by ensuring the fluency of replies, and the emotion is more prominent and delicate.
In the future, we will enhance the flexibility of the model by introducing other knowledge (such as a tone) and customize a personalized framework to meet the specific needs of the actual application.
\section*{Acknowledgment}
This work was supported by the State Key Program of the National Natural Science Foundation of China (61432004, 71571058, 61472117,and 61461045). This work was partially supported by a project funded by the China Postdoctoral Science Foundation (2017T100447). This research was also partially supported by a Qinghai Province Science and Technology Fund for fundamental and applied research (No. 2016-ZJ-743).

\ifCLASSOPTIONcaptionsoff
  \newpage
\fi

\begin{IEEEbiography}[{\includegraphics[width=1in,height=1.25in,clip,keepaspectratio]{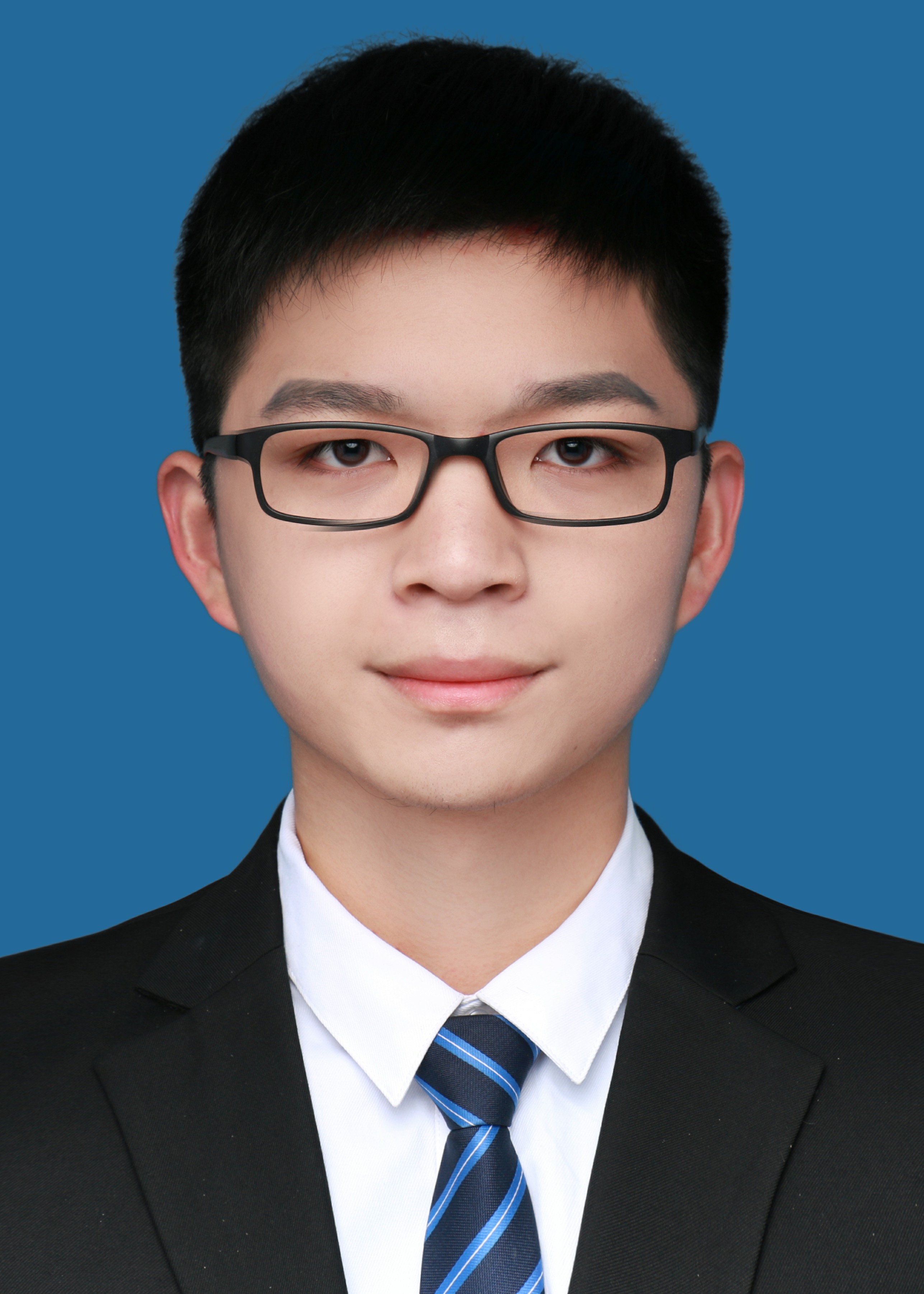}}]{Jia Li}
 	2016 undergraduate student, School of Computer and Information, Hefei University of Technology.The main research direction is natural language processing and emotional dialogue generation.
\end{IEEEbiography}

\begin{IEEEbiography}[{\includegraphics[width=1in,height=1.25in,clip,keepaspectratio]{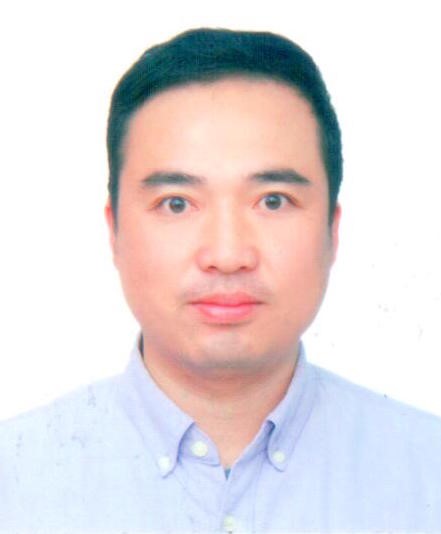}}]{Xiao Sun}
received his double doctor's degree in Dalian University of Technology(2010) of China and the University of Tokushima(2009) of Japan. He is now working as an associate professor in AnHui Province Key Laboratory of Affective Computing and Advanced Intelligent Machine at Hefei University of Technology. His research interests include Affective Computing, Natural Language Processing, Machine Learning.
Corresponding author of this paper(sunx@hfut.edu.cn) .
\end{IEEEbiography}

\begin{IEEEbiography}[{\includegraphics[width=1in,height=1.25in,clip,keepaspectratio]{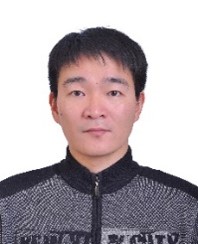}}]{Xing Wei}
Associate professor at Hefei University of Technology. His research interests include deep learning and Internet of things engineering, driverless solutions and so on.
\end{IEEEbiography}

\begin{IEEEbiography}[{\includegraphics[width=1in,height=1.25in,clip,keepaspectratio]{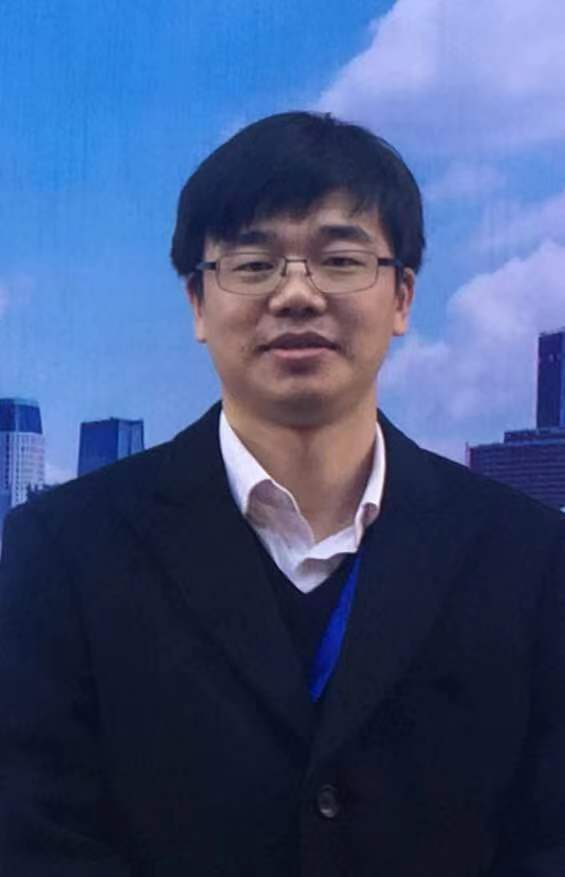}}]{Changliang Li}
Changliang Li received the Ph.D. degree from the Institute of Automation, Chinese Academy of Science, China, in 2015. Since 2018, he is currently the head with the Kingsoft institute of Artificial Intelligence. He has published widely in artificial intelligence and deep learning research. His current research interests include Deep Learning, Natural Language Processing and Data Mining.
\end{IEEEbiography}

\begin{IEEEbiography}[{\includegraphics[width=1in,height=1.25in,clip,keepaspectratio]{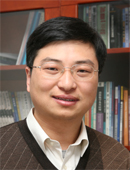}}]{Jianhua Tao}
Jianhua Tao received his M.S. degree from Nanjing University, Nanjing, China, in 1996, and his Ph.D. degree from Tsinghua University, Bei- jing, China, in 2001. He is currently a Professor with the National Laboratory of Pattern Recognition, Institute of Automation, Chinese Academy of Sciences, Beijing. His current research interests include speech synthesis and recognition, humancomputer interactions, and emotional information processing. He has authored over 60 papers in major journals and proceedings, such
as the IEEE TRANSACTIONS ON AUDIO, SPEECH, AND LANGUAGE PROCESSING, ICASSP, Interspeech, ICME, ICPR, ICCV, and ICIP. In 2006, he was elected as the Vice Chairperson of the ISCA Special Interest Group of Chinese Spoken Language Processing and an Executive Committee Member of the HUMAINE Association. He is an Editorial Board Member of the Journal on Multimodal User Interfaces and the International Journal of Synthetic Emotions, and a Steering Committee Member of the IEEE TRANSACTIONS ON AFFECTIVE COMPUTING.
\end{IEEEbiography}




\end{document}